\documentclass[letterpaper, conference]{ieeeconf}
\IEEEoverridecommandlockouts

\usepackage{amsmath,amssymb}
\usepackage{mathtools}
\mathtoolsset{showonlyrefs}
\usepackage{mathrsfs}
\usepackage{comment}
\usepackage[caption=false,font=footnotesize]{subfig}
\usepackage{multicol,multirow}
\usepackage[dvipsnames,table]{xcolor}
\usepackage{cite}
\usepackage{graphicx}

\newcommand{\spcmt}[1]{{\color{blue}\it (SP: #1)}}

\newcommand{\guidelines}[1]{{\color{TealBlue}{#1}}}

\title{\LARGE \bf Proactive Opinion-Driven Robot Navigation around Human Movers 
}

\author{Charlotte Cathcart, María Santos, Shinkyu Park, and Naomi Ehrich Leonard
\thanks{Cathcart, Santos, and Leonard are with the Department of Mechanical and Aerospace Engineering, Princeton University, Princeton, NJ 08544, USA. {\tt {\{cathcart, maria.santos, naomi\}@princeton.edu}}. Park is with the Electrical and Computer Engineering program, King Abdullah University of Science and Technology, Thuwal 23955, Saudi Arabia. {\tt {shinkyu.park}@kaust.edu.sa}.}
\thanks{This research has been supported by ONR grant N00014-19-1-2556, funding from KAUST, and Princeton University through the generosity of Lydia and William Addy '82.} \thanks{Study \#14788 has been approved by the International Review Board (IRB) of Princeton University. }}

\begin{document}

\maketitle
\thispagestyle{empty}
\pagestyle{empty}

\begin{abstract}
We propose, analyze, and experimentally verify a new proactive approach for robot social navigation driven by the robot's 
``opinion'' for which way and by how much to pass human movers crossing its path. 
The robot forms an opinion over time according to nonlinear  dynamics that depend on the robot's observations of human movers and its level of attention to these social cues. For these dynamics, it is guaranteed that when the robot's attention is greater than a critical value, deadlock in decision making is  broken, and the robot rapidly forms a strong opinion, passing each human mover even if the robot has no bias nor evidence for which way to pass.  
We  enable proactive rapid and reliable social navigation  by having the robot grow its attention across the critical value when a human mover approaches. 
With human-robot experiments we demonstrate the flexibility of our approach and validate our analytical results on  deadlock-breaking. 
We also show that a single design parameter can tune the trade-off between efficiency and reliability in human-robot passing. The new approach has the additional advantage that it does not rely on a predictive model of human behavior.

\end{abstract}


\section{Introduction}

\begin{figure}[t]

\centering
\subfloat[\label{subfig:multiple_humans_timelapse}]{\includegraphics[width=0.27\textwidth]{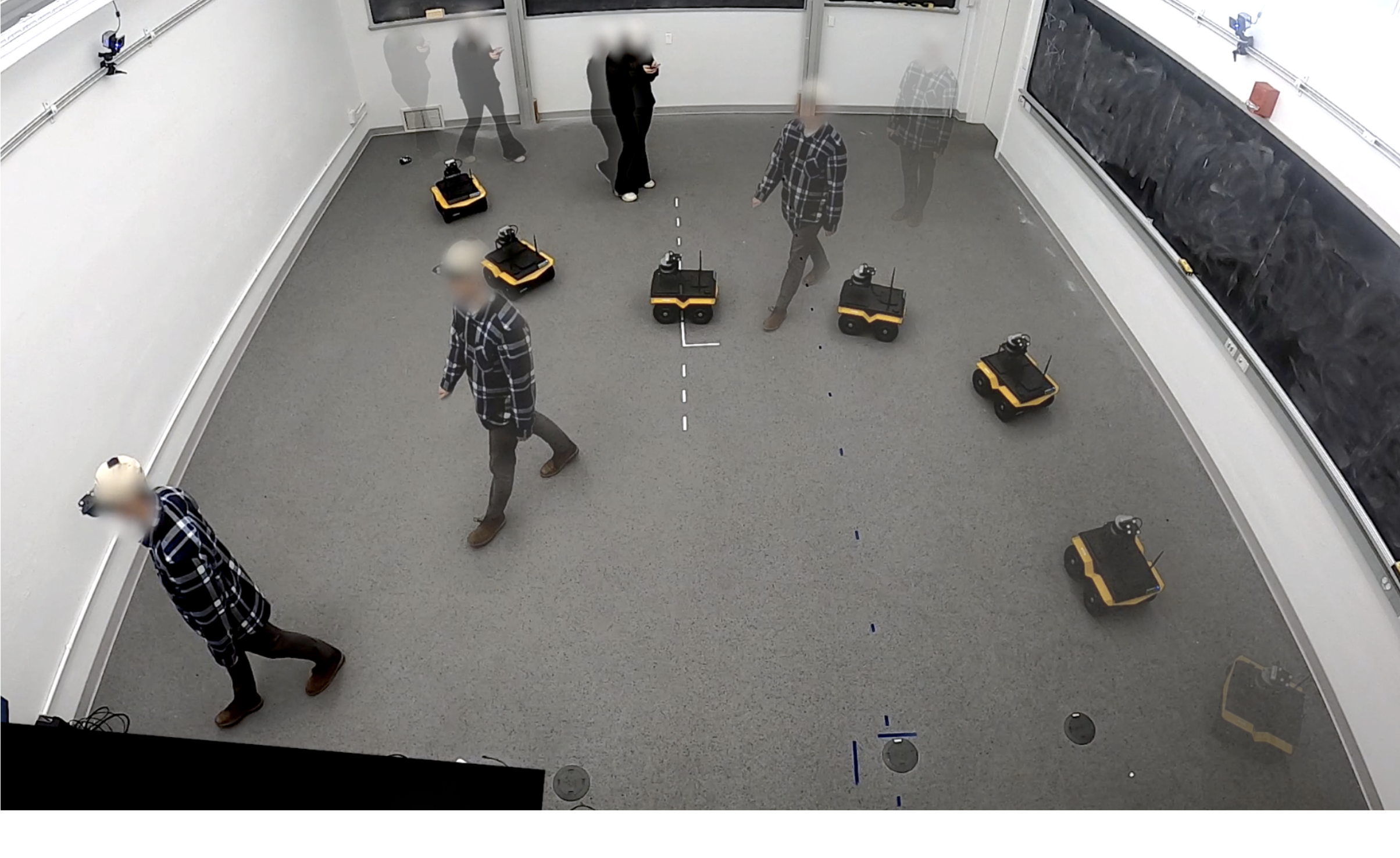}}
\subfloat[\label{subfig:multiple_humans_trajectories}]{\includegraphics[width=0.2\textwidth]{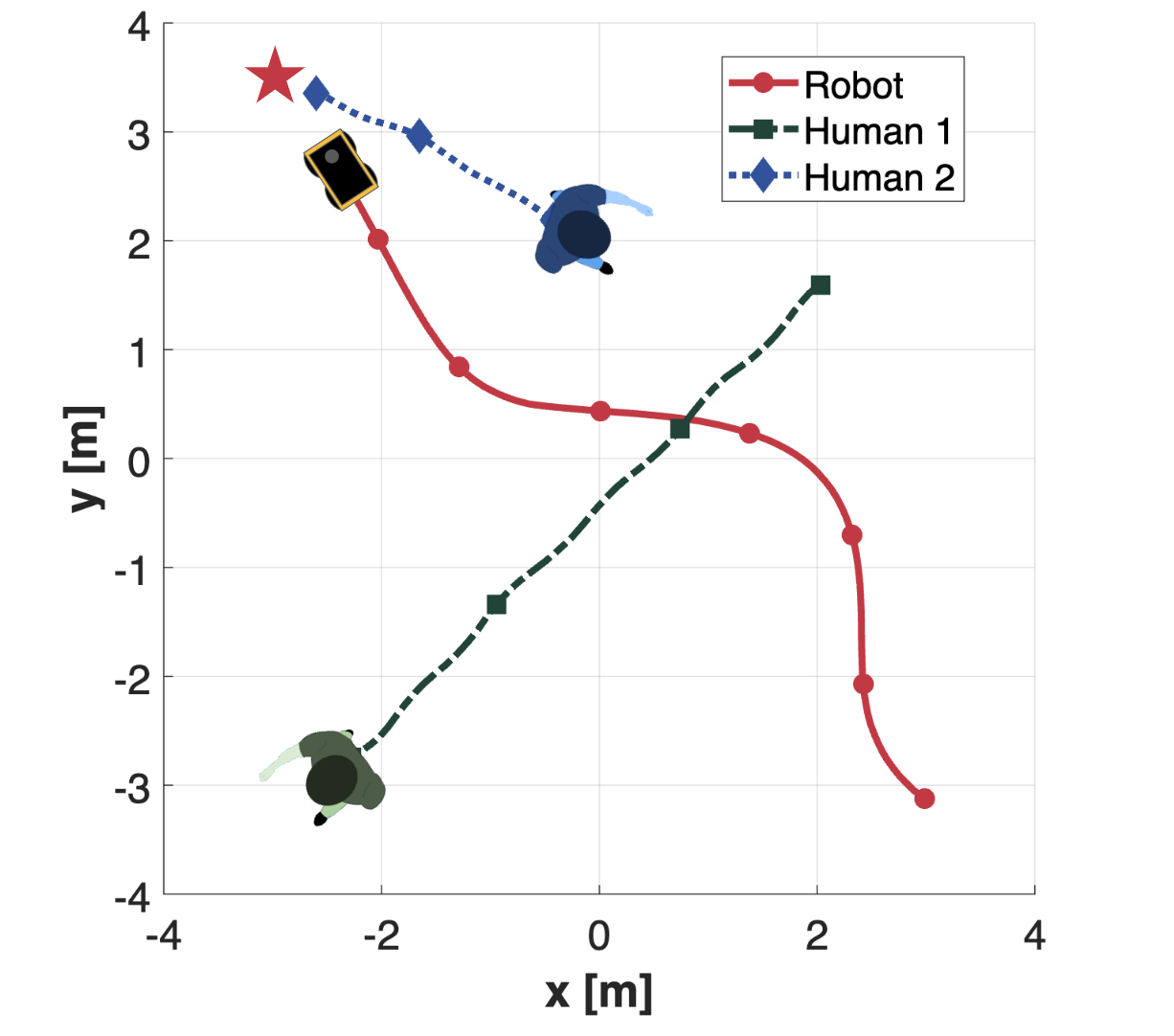}}


\caption{A robot using opinion-driven navigation to pass two humans. \protect\subref{subfig:multiple_humans_timelapse} Time-lapse of the experimental trial. \protect\subref{subfig:multiple_humans_trajectories} The full trajectories of the robot (red line) and two humans (blue and green lines) with temporal markers.}
\label{fig:passing_setup}
\end{figure}

Autonomous mobile robots are increasingly being used for tasks in settings such as warehouses and open public spaces where they will encounter human movers.
In order to accomplish their tasks in these settings,  the  robots need to reliably and gracefully navigate around human movers. 
In this paper, we propose, analyze, and experimentally verify a new approach for the social navigation of a mobile robot.
Fig.~\ref{fig:passing_setup} shows experimental results of a mobile robot navigating around two human movers using the new approach.

We build on  the nonlinear opinion dynamics model presented in  \cite{ref:Bizyaeva2022_NODM_with_Tunable_Sensitivity} and propose an approach that allows a robot to rapidly form an \textit{opinion} that represents the strength of its preference for which direction---left or right---it will use to pass each human mover crossing its path. This opinion, in turn, drives the robot's motion, modifying its nominal path to reliably pass the human. A key to the opinion dynamics is that when the robot's \textit{attention} to social cues grows above a critical value, the neutral opinion to stay the course is destabilized and the robot rapidly forms a strong and stable opinion for moving in one of the two passing directions. Our approach is therefore to design dynamics for the robot's attention that drive it above this critical value when the robot senses a human mover approaching its path. The active control of attention yields a rapid and reliable passing motion in response to an approaching human mover; 
this renders our approach ``proactive'' rather than merely ``reactive.'' 

\begin{figure*}
\centering
\subfloat[\label{subfig:two_humans_ZU1}]{\includegraphics[width=0.16\textwidth]{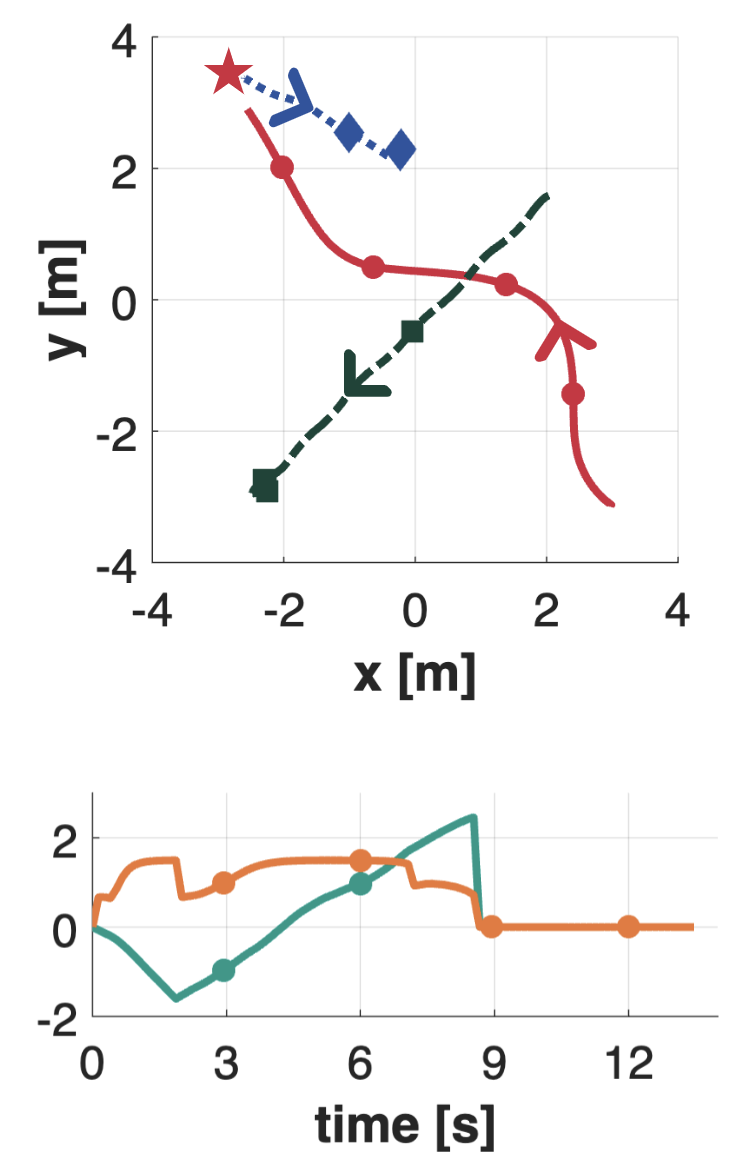}}
\hspace{1.3em}
\subfloat[\label{subfig:two_humans_ZU2}]{\includegraphics[width=0.16\textwidth]{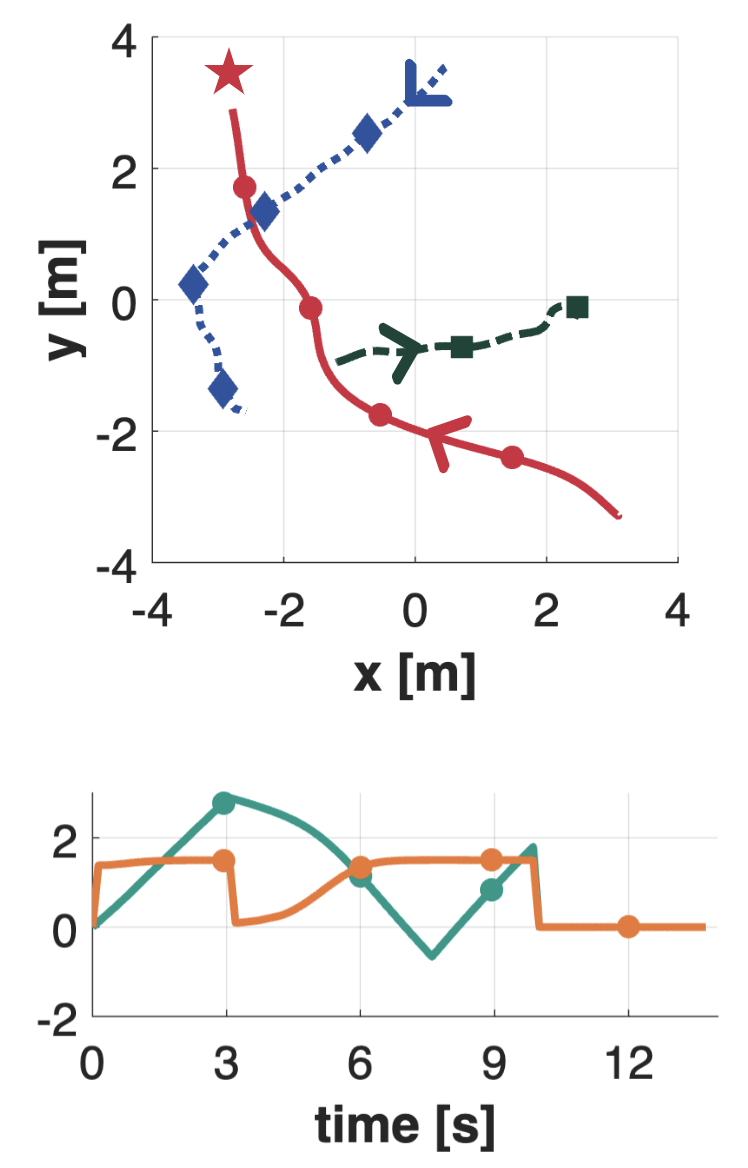}}
\hspace{1.3em}
\subfloat[\label{subfig:two_humans_ZU3}]{\includegraphics[width=0.16\textwidth]{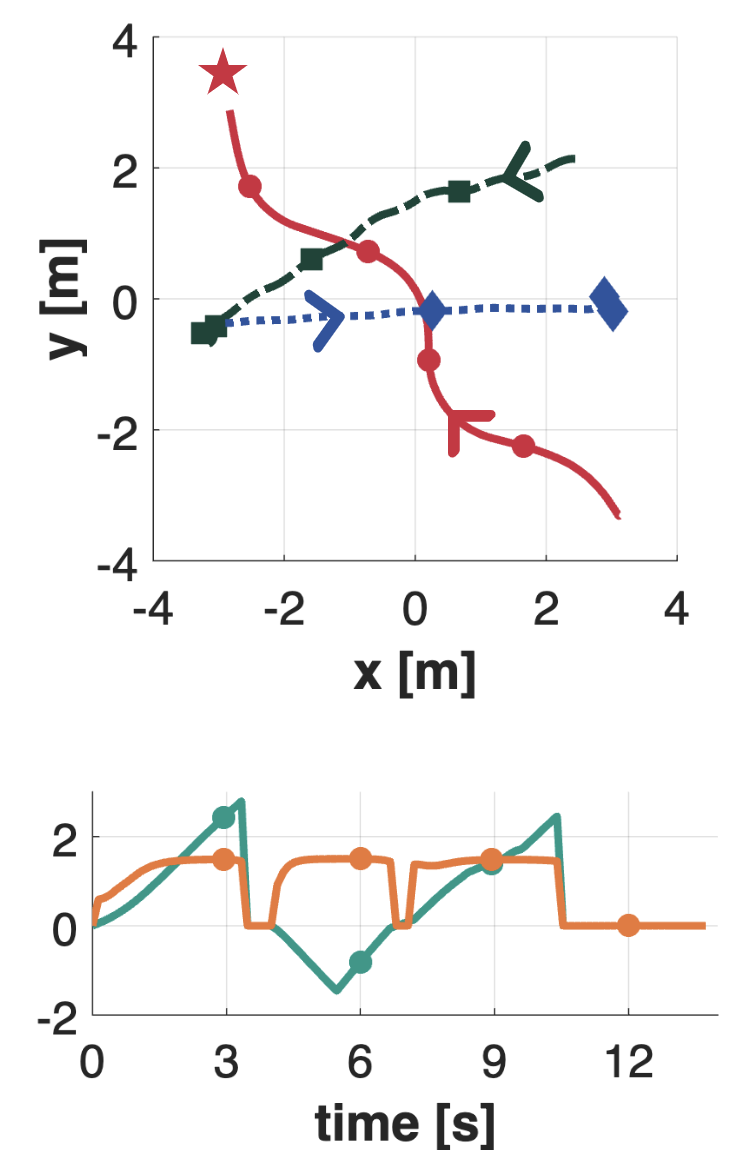}}
\hspace{1.3em}
\subfloat[\label{subfig:two_humans_ZU4}]{\includegraphics[width=0.16\textwidth]{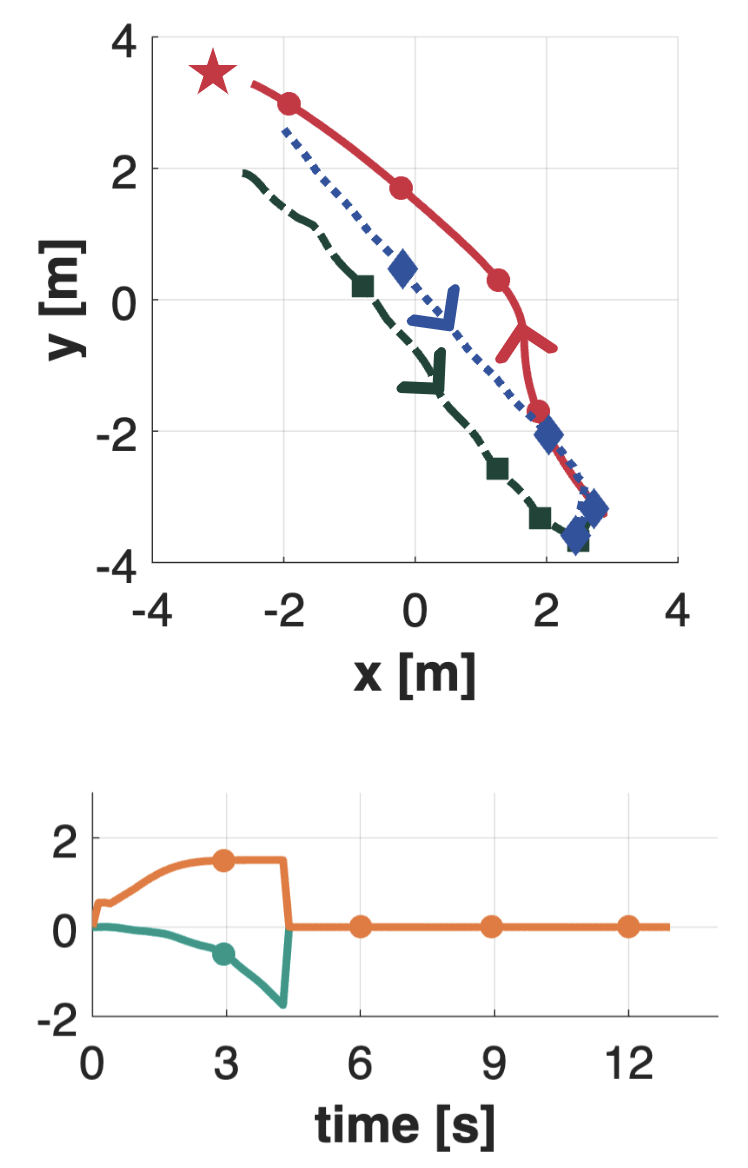}}
\hspace{1.3em}
\subfloat[\label{subfig:two_humans_ZU5}]{\includegraphics[width=0.16\textwidth]{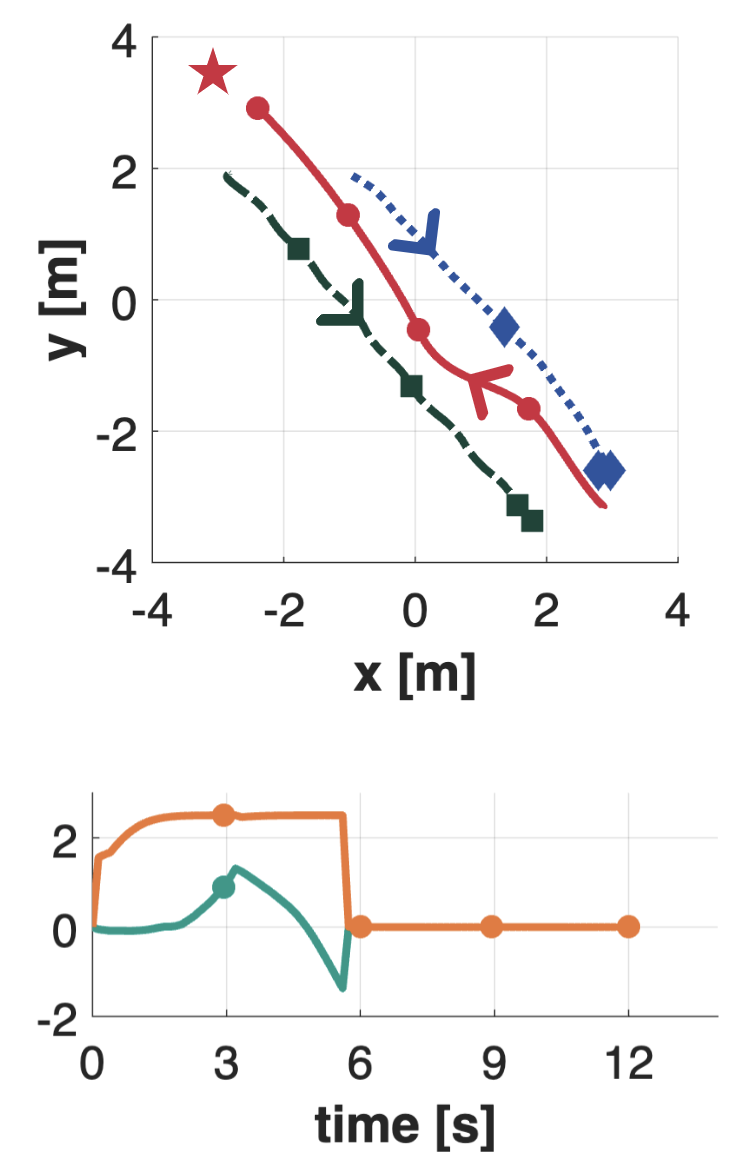}}\\
\subfloat{\includegraphics[width=0.88\textwidth]{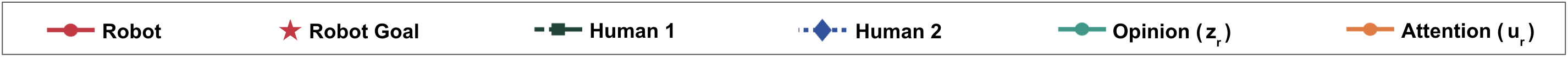}}
\caption{Multiple experimental trials with two humans and a robot using the new approach. The top row shows the complete trajectories of the robot (red line) and humans (green and blue lines) over the course of a trial as the robot moves toward its goal (red star). Each trajectory is marked with an arrow indicating the mover's direction. The bottom row shows the robot's opinion $z_r$ (teal line) and attention $u_r$ (orange line) over the course of the trial above it. Temporal markers (dots) are shown along spatial trajectories, opinion, and attention. See Section~\ref{sec:multi_human_experiments} for parameters used.} 
\label{fig:two_humans_experiment}
\end{figure*}

Once the robot passes a human, its opinion with respect to that human is no longer relevant; the opinion quickly returns to its neutral value, allowing the robot to continue towards its destination. Likewise, the robot's attention also goes to zero, making the robot ready for new potential conflicts with other movers. Figs. \ref{fig:passing_setup} and \ref{fig:two_humans_experiment} provide experimental results of the robot  navigating different encounters when traveling to a goal destination that is diagonally across an open space with two humans moving and pausing in a variety of scenarios.

Opinion dynamics are used to enable decision making in multi-agent systems in a range of tasks 
\cite{ref:Hamann2018_Opinion_Dynamics_Mobile_Agents, ref:MontesdeOca_2010_Opinion_Dynamics_Decentralized_Decision_Making, ref:Bizyaeva2022_Decentralized_Control_Switching}.
In the nonlinear opinion dynamics of \cite{ref:Bizyaeva2022_NODM_with_Tunable_Sensitivity},
an agent's opinion 
is influenced by the opinions of others when its 
attention exceeds a critical level. 
At this point the agents are guaranteed to form strong opinions (e.g., to agree on or coordinate among options), 
hence avoiding indecision, i.e., deadlock in their decision making. 
In the robot social navigation problem, we leverage the deadlock breaking guarantees of 
the coupled attention-opinion dynamics to ensure that, when necessary to avoid an approaching human mover, the robot will  rapidly select and move in one of the two passing directions even if there is no indication from the human or the environment that one direction is better than the other, or if the robot's bias for one direction or the other, if it has one, conflicts with the human's chosen passing direction.

Of relevance to our work is the literature on robot social navigation (see recent survey articles \cite{Magrogiannis2023_core_challenges,ref:Stone2021_Social_Navigation_Survey, ref:Gao2021_Socially_Aware_Nav_Survey,ref:Hart_HumanSignals2020,ref:Mavrogiannis2022_Social_Momentum} and references therein), where a common theme is in investigating the design of navigation algorithms for autonomous robots to safely and comfortably interact with the humans they encounter. Earlier work \cite{ref:Helbing1995_Social_Force_Model} in modeling human navigation behavior proposes a model based on the observation that the motion of pedestrians is subject to \textit{social forces}. More recent works \cite{ref:Reddy2021_Social_Cues_as_Forces, ref:Kivrak2021_Human_Inhabited_environments} incorporate \textit{social cues} into 
the social force model 
and the improved models are used to design robot navigation algorithms. The work of \cite{ref:Kirby2009_Companion} proposes a constrained optimization approach to design a navigation algorithm that penalizes the robot when its behavior violates conventions observed in the human's navigation. In \cite{ref:Mavrogiannis2022_Social_Momentum}, a reactive control policy is used to follow and maintain the passing sides observed by passing humans through social momentum.  
References such as \cite{ref:Samsani2021_Socially_Compliant_Robot_Nav_using_RL, 8202312, ref:Kollmitz2020_Learning_from_Physical_Interaction_viaIRL} explain learning-based approaches that leverage the recent advancement in deep reinforcement learning to train mobile robots through multiple trial-and-error processes to safely navigate in human-populated areas.

Another important line of research in the social navigation literature is data-driven learning approaches that infer human navigation models from their demonstration data, and use the models to predict human motions and to design robot motion planners.
The work of \cite{ref:Bera2017_SocioSense} leverages Bayesian learning to construct a motion model and personality characteristics of pedestrians, and use predicted pedestrian trajectories from the model for socially-aware robot navigation. 
Inverse Reinforcement Learning (IRL)-based approaches, for instance \cite{ref:Kretzschmar2016_IRL_Socially_Compliant_Robot, ref:Kollmitz2020_Learning_from_Physical_Interaction_viaIRL, ref:Okal2016_Learning_SocialNorm_RobotNav_BayesianIRL}, take human demonstration data to estimate a utility function used in human navigation tasks, and use it to generate robot trajectories that imitate the demonstrated human motions. In particular, a recent relevant work \cite{ref:Che2020_Social_Nav_Explicit_Implicit_Communication} studies the effect of human-robot communication in social navigation and proposes an IRL-based robot planning framework to generate communication actions that maximize the robot's transparency and efficiency.

Our work is distinct in that 1) it is proactive rather than reactive, 2) it does not require constructing a predictive model of human navigation as in IRL-based approaches, rather it only needs the robot to observe the position and moving direction of the human, and 3) our robot navigation model is analytically tractable so that we can establish a guarantee on deadlock-free decision making in the robot-human navigation.
This contrasts with the reinforcement learning approaches, which are in general difficult to analyze, and existing reactive approaches, such as social force models, which do not provide the same deadlock-free guarantee.

In Section~\ref{section:social_navi}, we introduce the nonlinear opinion dynamics and propose a new model for robot navigation in a human-robot navigation setting. In Section~\ref{section:analysis}, using tools from nonlinear dynamical systems theory, we discuss how the model ensures rapid deadlock-free robot navigation. To demonstrate and test the flexibility of our approach, we carry out experiments with  two human participants and a mobile robot in a range of scenarios, which we report on in Section~\ref{sec:multi_human_experiments}. We examine and validate the effectiveness of rapid deadlock-free navigation with further experiments in Section ~\ref{sec:one_human_experiments}. We conclude with a discussion in Section~\ref{section:conclusion}.

\section{Nonlinear Opinion Dynamics \\in Social Navigation} \label{section:social_navi}

We study a robot navigation problem where a robot approaches and passes human movers while traveling to its destination (see examples in Figs.~\ref{fig:passing_setup} and \ref{fig:two_humans_experiment}). In this context, we want to enable the robot to repeatedly overcome human movers in a rapid and reliable fashion. We are also interested in tackling challenging scenarios such as the human-corridor passing problem \cite{ref:Pacchierotti2005_Hallway_Setting_HRI_Study, ref:Thomas2018_Doorway_Negotiation, ref:Lu2013_Corridor} that may result in deadlock if, for example, both the robot and the human have conflicting passing biases.
In these situations, a key objective 
is to ensure 
that the robot moves reliably around the human regardless of the human's awareness of the robot. It is also desirable that the robot moves efficiently around the human. However, reliability and efficiency are in tension: giving the human a lot of space may create reliably successful but inefficient passing 
whereas giving the human only a little space is efficient but creates less reliably successful passing. 

To address these competing objectives, we propose a new dynamic model for robot navigation
based on the nonlinear opinion dynamics of \cite{ref:Bizyaeva2022_NODM_with_Tunable_Sensitivity}. 
We review these dynamics in Section~\ref{sec:NODM}. We specialize the dynamics to proactive opinion-driven robotic navigation in Section~\ref{sec:robotmodel} and show how a single design parameter can be used to control the reliability-efficiency trade-off. In Section~\ref{section:analysis}, we provide analysis that shows how deadlock breaking is guaranteed. 


\subsection{Nonlinear Opinion Dynamics Model\label{sec:NODM}}

Consider a system of $N_a$ agents forming opinions about two options.
Let $z_i \in \mathbb{R}$ be the opinion of agent $i$, which represents the strength of its preference for  option 1 if $z_i > 0$ and for option 2 if $z_i < 0$. It is indifferent, i.e., neutral, if $z_i = 0$. Strength of preference is $|z_i|$.
The nonlinear opinion dynamics model, described below, explains how each agent~$i$ updates its opinion $z_i$ continuously over time in response to its own opinion, the opinions of others $z_k$, and any internal bias or external stimulus $b_i$. Letting $\dot z_i = dz_i/dt$,
\begin{equation} \label{eq:general_opinion_model}
\dot{z}_i = -d_i \, z_i + u_i \tanh{ \Big(\alpha_i z_i + \gamma_i \textstyle\sum_{\substack{k \neq i \\ k=1}}^{N_a} a_{ik} z_k + b_i\Big)}.
\end{equation}

The opinion $z_i$ can be interpreted as the discounted accumulation of \textit{social influence} weighted by the parameter $u_i \geq 0$. 
The social influence is defined as the hyperbolic tangent function of the weighted sum of the opinion $z_k$ of every agent $k$ observed by agent $i$ 
and a bias/stimulus $b_i$.
The \textit{resistance parameter} $d_i>0$ defines the rate of exponential discount in the accumulation of the social influence.
The \textit{attention} $u_i \geq 0$ is a tuning variable, which can be adjusted to reflect the agent's (changing) effort to pay attention to the social influence. The parameter $a_{ik}=1$ if agent $i$ can observe agent $k$; otherwise, $a_{ik} = 0$. The parameters $\alpha_i>0$ and $\gamma_i \in \mathbb{R}$ are weights defining how much influence $z_i$ and $z_k$, respectively, have on agent $i$'s opinion update. If $b_i > 0$ (resp.\ $b_i < 0$), the bias is for option 1 (resp.\ for option 2). In case of no bias, we set $b_i = 0$. 
\subsection{Dynamic Model for Opinion-Driven Robot Navigation} \label{sec:robotmodel}
\begin{figure}[t]
\centering
\includegraphics[width=0.83\linewidth]{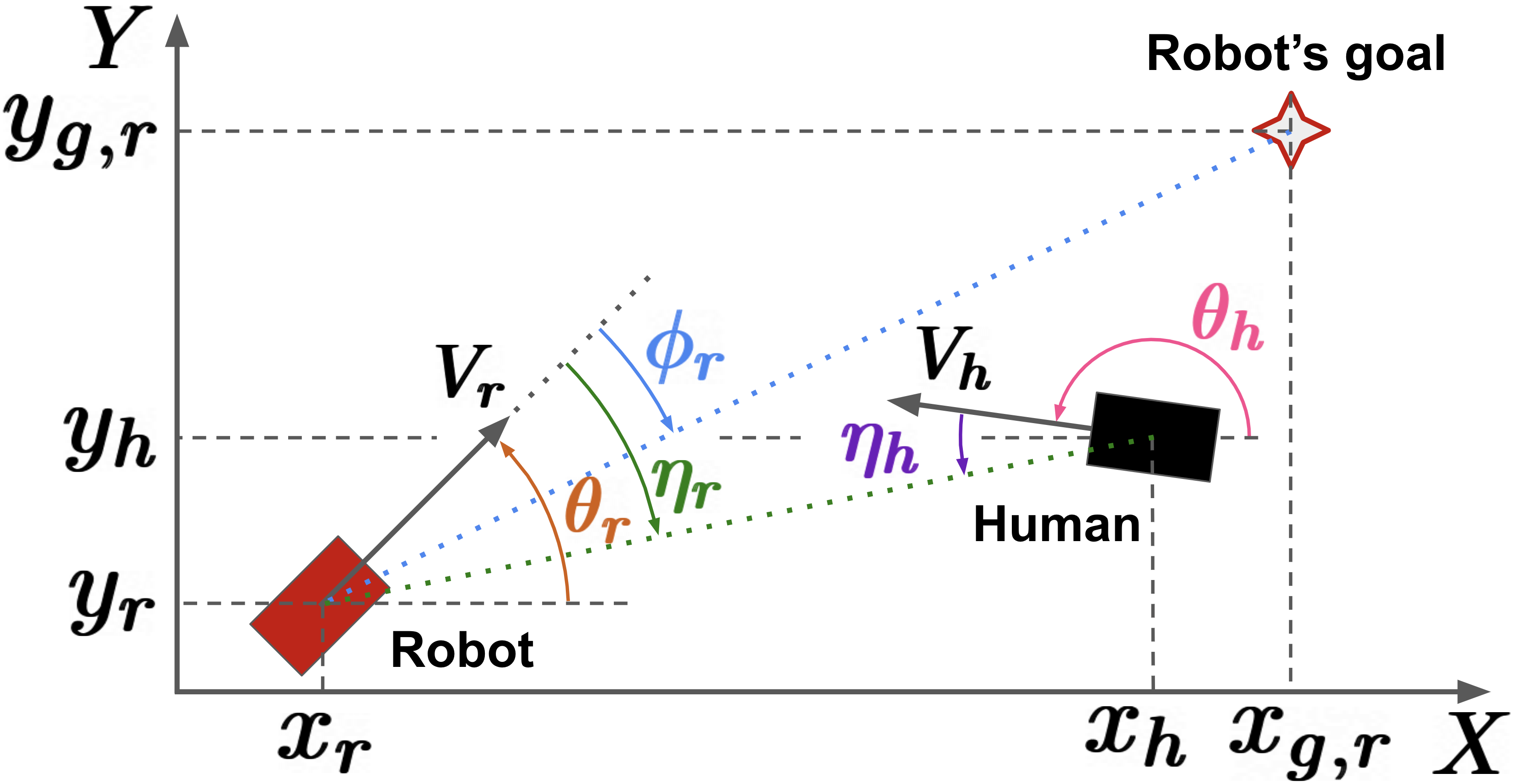}
\caption{An illustration of notation for human-robot passing. 
}
\label{fig:rot_human_geometry_setup}
\end{figure}

Building on \eqref{eq:general_opinion_model}, we propose a robot navigation model that forms an opinion to drive the robot's motor control in an uncrowded and uncluttered environment with human movers. We assume that the robot moves at a constant speed $V_r$, but can regulate its angular velocity. We represent the robot's position and heading angle as $\mathbf{x}_r = (x_r, y_r)$ and $\theta_r$, respectively. 
For each human $j$ that the robot can detect, we denote their speed  $V_{h_j}$, position $\mathbf{x}_{h_j} = (x_{h_j}, y_{h_j})$, and heading angle $\theta_{h_j}$. Let $\eta_{r_j}$ be the heading of the robot relative to the line between the robot and the human $j$. Let  $\eta_{h_j}$ be the heading of  human $j$ relative to the line between the robot and the human. See Fig.~\ref{fig:rot_human_geometry_setup} for illustration of notation.

The robot focuses on  
the human mover $j$ that minimizes $\chi_j/\kappa_j$ where $\chi_j = \| \mathbf{x}_{r} - \mathbf{x}_{h_j}\|$, $\kappa_j = \cos \eta_{h_j}$,
and $\eta_{h_j} \in (-\frac{\pi}{2},\frac{\pi}{2})$.  
This is the human who is most rapidly approaching the robot. 
We use $\mathbf{x}_h(t)$, $\eta_h(t)$, $\chi(t)$, and $\kappa(t)$, i.e., without index $j$, 
to refer to whichever human is the one most rapidly approaching the robot at time $t$. 

We define $z_r > 0$ (resp.\ $z_r < 0$) as the robot's strength of preference for moving left (resp. right). When $z_r = 0$, the robot's opinion is neutral, i.e., it is indifferent to these options. Our approach does not require any knowledge of a human model; however, we assume that the robot can measure $\eta_h$ and use it as a proxy for the robot's perception of the human's opinion on direction  as $\hat{z}_h = \tan \eta_h$\footnote{We resort to \cite{ref:Fiore2013_Robot_Gaze_Proxemic_Behavior, ref:Unhelkar2015_HRI_Anticipatory_Indicators, ref:Ratsamee2015_Face_Orientation_Human_Path_Prediction} for the basis for estimating the human's navigation intent using their orientation.}. This is unlike other approaches that require a longer-term prediction of human trajectories, such as \cite{ref:Kretzschmar2016_IRL_Socially_Compliant_Robot, ref:Kollmitz2020_Learning_from_Physical_Interaction_viaIRL, ref:Okal2016_Learning_SocialNorm_RobotNav_BayesianIRL, ref:Che2020_Social_Nav_Explicit_Implicit_Communication}.



Our proactive opinion-driven robot navigation model specifies (a) how the robot's opinion $z_r$ changes in response to its attention $u_r$, its current opinion, its estimate $\hat{z}_h$ of the opinion of the focal human mover, and possibly a bias $b_r$;  (b) how the robot's attention $u_r$ changes in response to $\kappa$ and $\chi$; and (c) how the robot's heading $\theta_r$ changes as a function of its opinion $z_r$ and the direction $\phi_r$ to its goal:
\begin{subequations} \label{eq:navi_dynamics}
\begin{align}
 \dot{z}_r &= -d_r \, z_r + u_r \tanh\left( \alpha_r z_r + \gamma_r \hat{z}_h + b_r \right), \label{eq:full_z_r} \\
 \tau_u \dot{u}_r &= -u_r + g(\kappa, \chi; R_r), \label{eq:full_u_r} \\
 \dot{\theta}_r &= k_{r} \sin \left( \beta_r \tanh z_r + \phi_r \right), \label{eq:full_theta_r}
\end{align}
\end{subequations}
where $d_r$, $\alpha_r$, $\gamma_r$, $\tau_u$, $R_r$, $k_r > 0$ and $\beta_r \in (0, \frac{\pi}{2}]$ are design parameters. 
Note that \eqref{eq:full_z_r} is similar to \eqref{eq:general_opinion_model} except the human's opinion $z_h$ is replaced with the proxy $\hat{z}_h = \tan \eta_h$. 

 We design the attention dynamics \eqref{eq:full_u_r} so $u_r$ grows quickly  
 when a human mover gets close. Unless otherwise noted, we let $\tau_u \rightarrow 0$ and define $g$ using a Hill function to get 
\begin{equation}
\label{eq:u_r_Hillfxn}
    u_r = g(\kappa, \chi; R_r) = \underline{u} + (\bar{u} - \underline{u}) \left(\frac{(R_r \kappa)^n} {(R_r \kappa)^n + \chi^n }\right),
\end{equation}
 where $0 \leq \underline{u} < \bar{u}$ and $n > 0$. 
 The variable $u_r$ increases from $\underline{u}$ as the robot and human move closer towards collision, based on a critical distance parameter $R_r > 0$, and saturates at the value $\bar{u}$. This drives $u_r$ above a critical value that destabilizes the neutral opinion $z_r = 0$, allowing the robot to rapidly form a strong opinion  when a human mover approaches, and thus rapidly pass the human on one side or the other. In this sense our approach is proactive. See Section~\ref{section:analysis} for a rigorous analysis of the deadlock breaking.  


To understand the role of design parameter $\beta_r \in (0, \frac{\pi}{2}]$, note that when $z_r$ is sufficiently large so that $\tanh z_r \approx 1$ (resp. $-1$), \eqref{eq:full_theta_r} steers the robot's heading angle an additional $\beta_r~\mathrm{radians}$ in the counterclockwise (resp. clockwise) direction from the orientation to the goal location. Hence, we can tune $\beta_r$ to prescribe how much the robot's heading angle should deviate from its direct path to its goal when it detects the human and forms a strong opinion on its passing direction. In this way the parameter $\beta_r$ can be used to tune the reliability-efficiency trade-off as we show through the deadlock breaking human-robot experiments described in Section~\ref{sec:one_human_experiments}.



\label{sec:modeldesign}
Our approach can be extended to incorporate path planning, e.g., to avoid driving the robot to a local minimum in the case of a cluttered environment. 
For example, this would be possible using a path planning approach such as the rapidly-exploring random tree (RRT) in place of \eqref{eq:full_theta_r}, 
with opinion $z_r$ as an input.
This would regulate not only the robot's angular velocity but also its moving speed. 

\section{Guarantee on Deadlock-Free Navigation} \label{section:analysis}
A key contribution of our work is in guaranteeing deadlock-free navigation. We establish such a performance guarantee by analyzing the robot navigation model \eqref{eq:navi_dynamics}
. In particular, we discuss how the robot can rapidly and reliably form a strong opinion to select one of the two options---move left ($z_r>0$) or right ($z_r<0$)---and avoid colliding with a human, even when the human maintains a path straight for the robot and the robot has no bias ($b_r=0$) on which way to pass. To establish this, we use tools from nonlinear dynamical systems theory \cite{ref:Guckenheimer2002} to show that there is a deadlock-breaking \textit{pitchfork bifurcation} in \eqref{eq:navi_dynamics} when the robot's attention $u_r$ reaches a critical level $u_r^*$ (as it nears the human), corresponding to the destabilizing of the deadlock solution and the emergence of bi-stable solutions for moving left and for moving right. 

We examine the challenging case in which the human  does not react to the robot's movement. 
We validate our analysis through human-robot experiments in Section~\ref{sec:one_human_experiments}.

\begin{figure}[t]
  \centering
  \begin{minipage}[b]{\linewidth}
  \setlength{\parindent}{1.5em}
  \begin{minipage}[b]{.49\linewidth}
    \centering
    \subfloat[Symmetric pitchfork.]{\label{subfig:symmetric_pitchfork}\includegraphics[width=\linewidth]{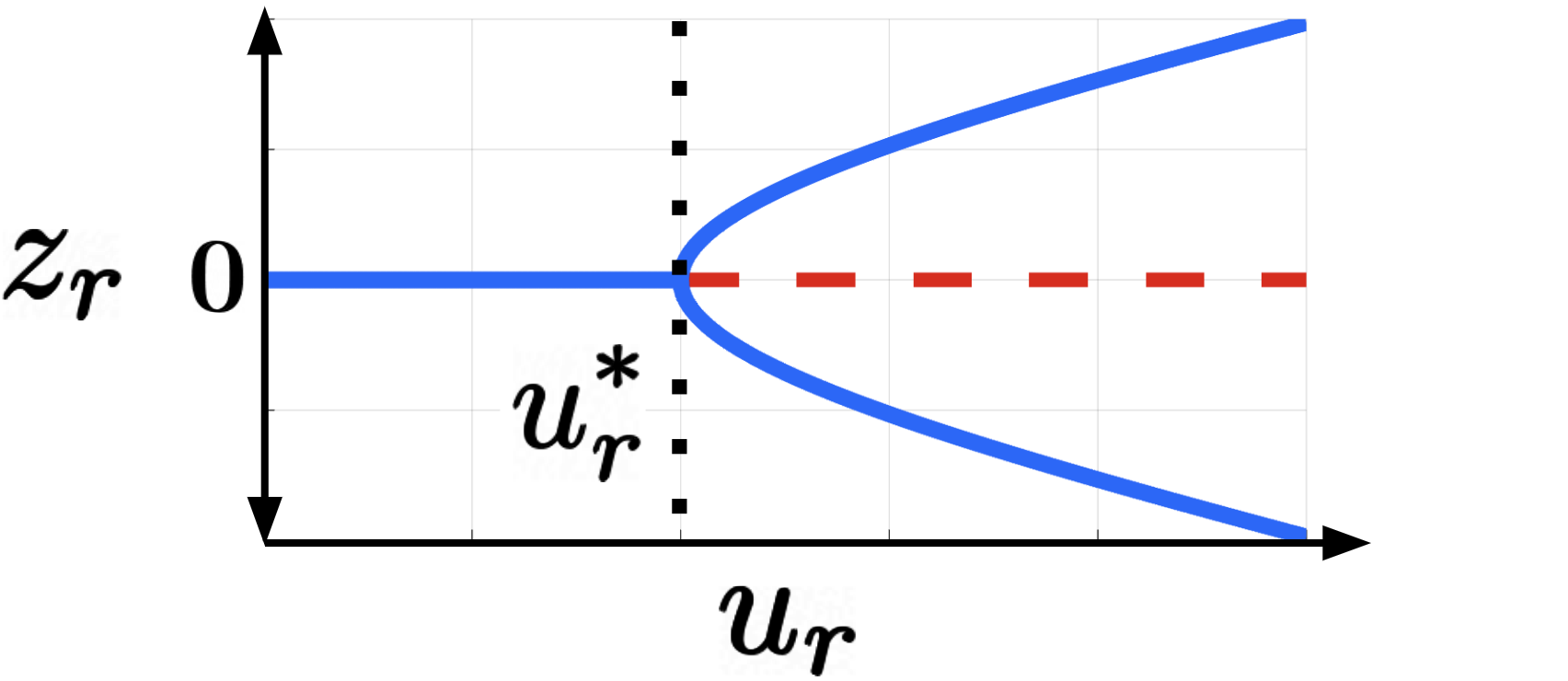}}\\
    \subfloat[Unfolded pitchfork.]{\label{subfig:unfolded_pitchfork}\includegraphics[width=\linewidth]{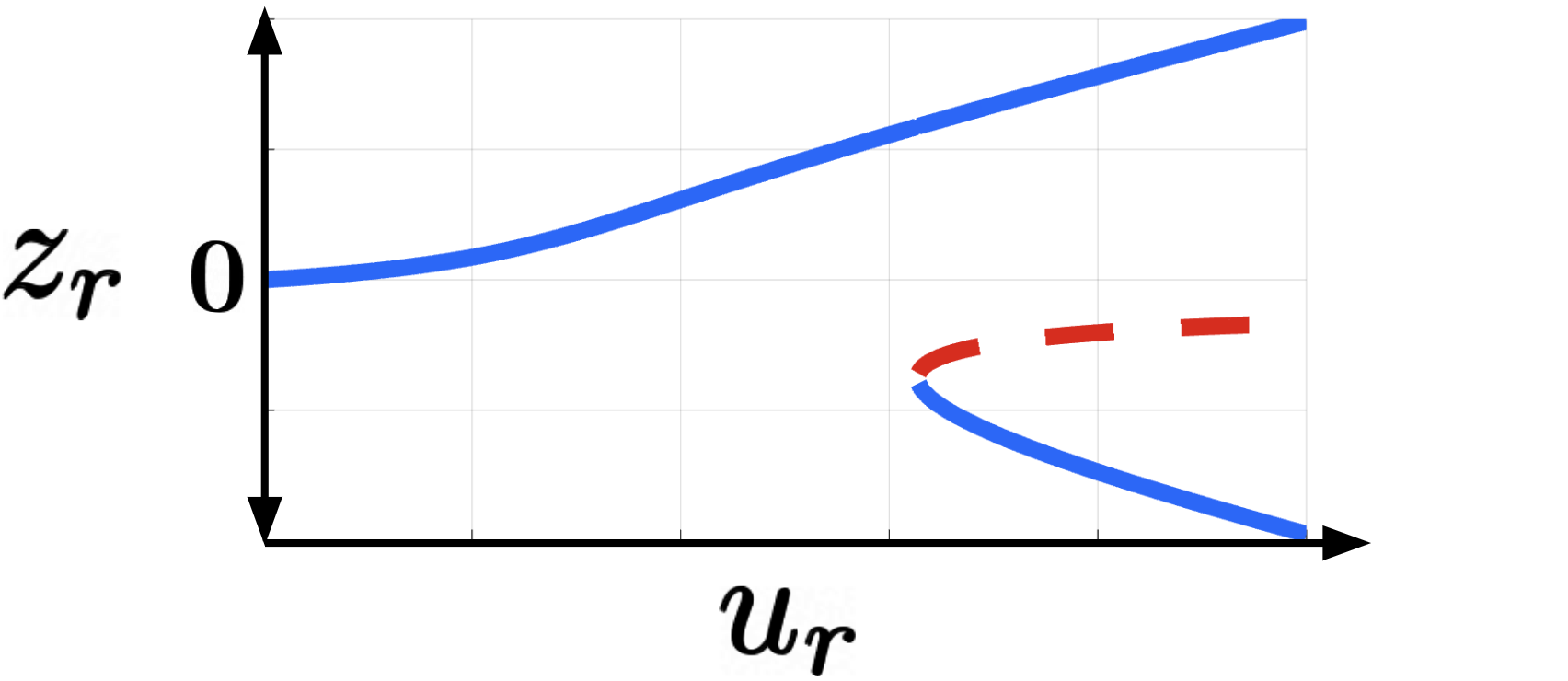}}\\
  \end{minipage}%
  \begin{minipage}[b]{.34\linewidth}
    \centering
    \setlength{\parindent}{0.5em}
    \subfloat[Simulation.]{\label{subfig:unaware_human_simulations}\includegraphics[width=\linewidth]{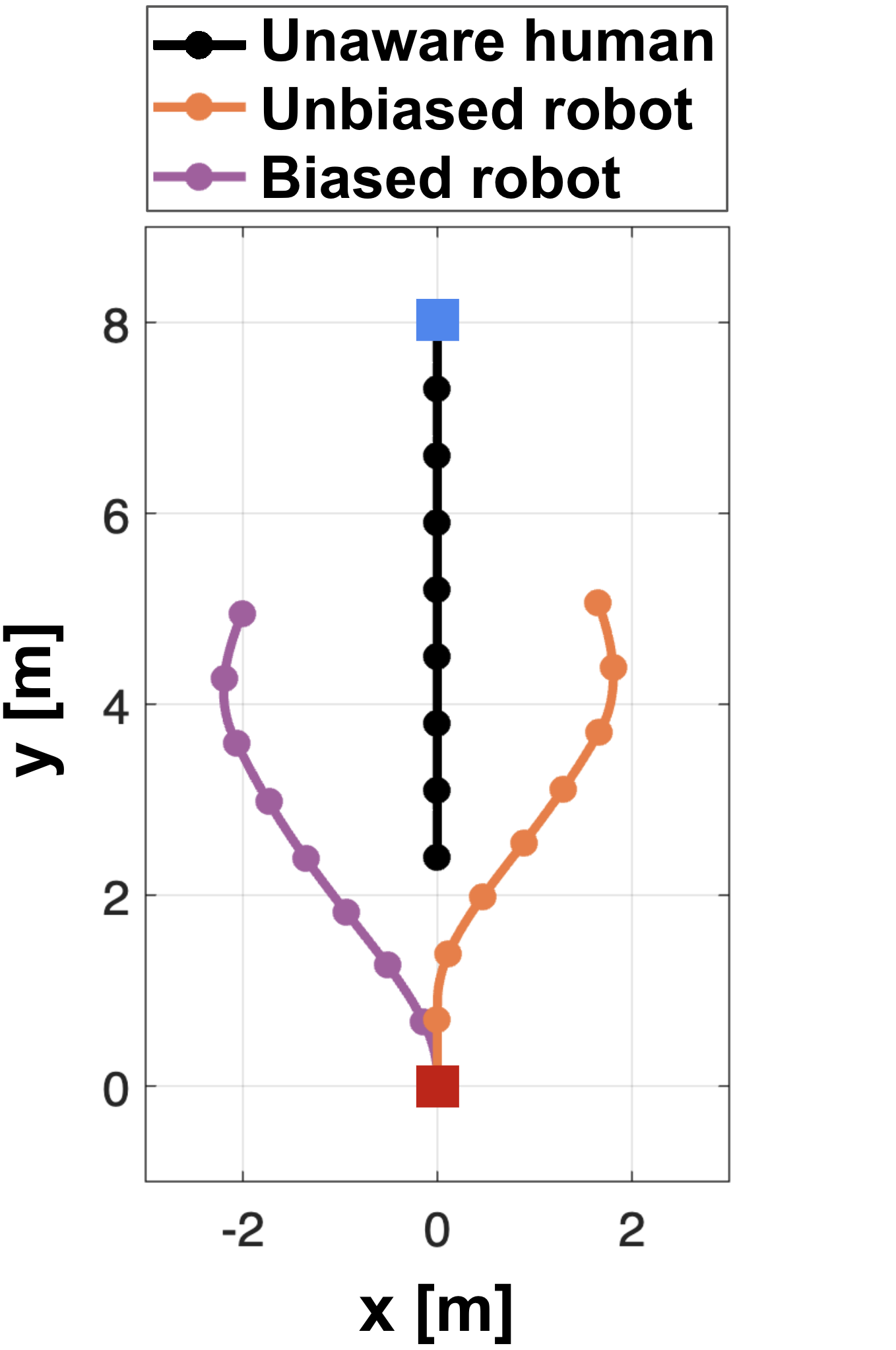}}
    \end{minipage}\par\medskip
    \end{minipage}
  \caption{Analysis of deadlock breaking in the robot's opinion dynamics when the human moves straight towards the robot. \protect\subref{subfig:symmetric_pitchfork} When the robot is unbiased $(b_r=0)$, deadlock is broken as $u_r$ increases above critical value $u_r^*$, where two stable (blue solid) symmetric opinionated solutions emerge and deadlock becomes unstable (red dashed). \protect\subref{subfig:unfolded_pitchfork} When the robot is biased ($b_r =0.5$), the bifurcation ``unfolds'' where deadlock breaks but the likelihood of converging on one opinionated solution is greater than on the other. \protect\subref{subfig:unaware_human_simulations} Simulations of social navigation dynamics. Initial conditions for the robot and human indicated with red and blue boxes. Parameters of \eqref{eq:navi_dynamics}: $d_r = \alpha_r = 0.1$, $\gamma_r = 3$, $\tau_u = 1$, $g(\kappa, \chi; R_r) =\exp(\kappa (R_r - \chi))$ with $R_r = 16$,  $k_{r} = 1$, and $\beta_r=\pi/4$. }
  \label{fig:pitchforks_and_simulations}
\end{figure}

\begin{figure*}
\centering
\includegraphics[width=0.87\linewidth]{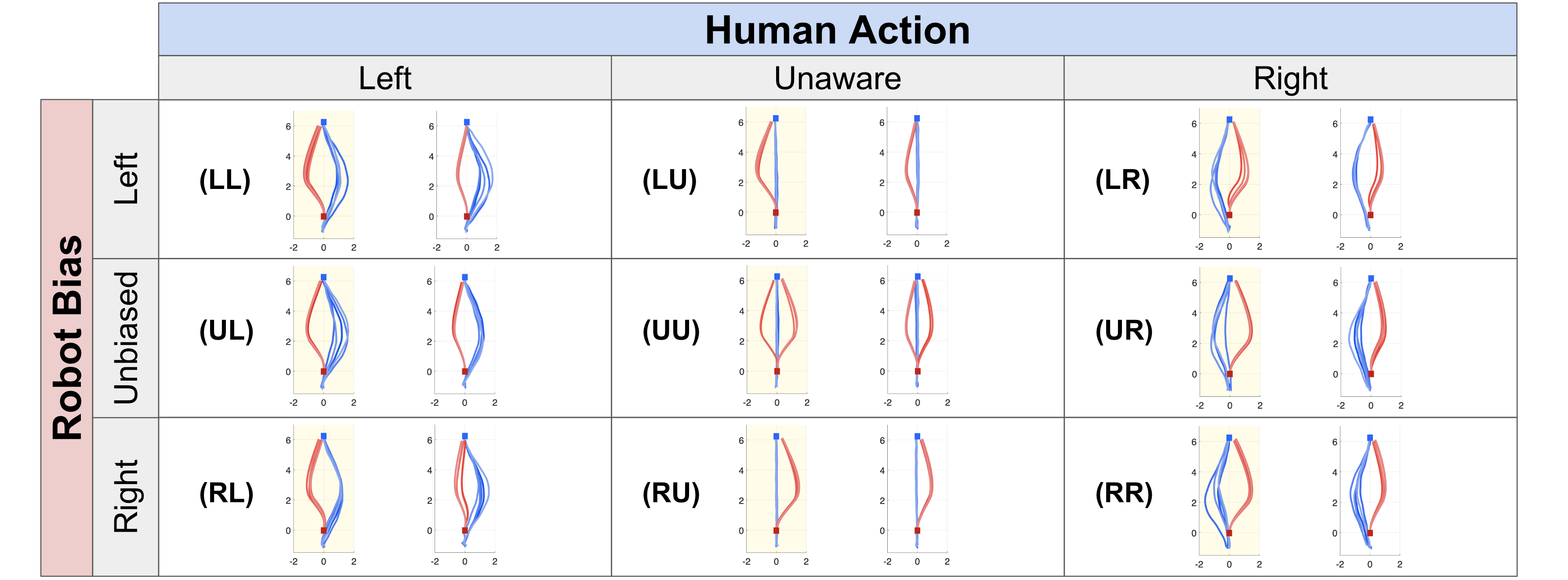}
\caption{The trajectory data for five runs each of the nine trial configurations for the case $\beta_r = \pi/4$ (shaded yellow) and for the case $\beta_r = \pi/6$ (unshaded). Axes correspond to the $xy$-plane in meters. The robot paths are shown in red with a red box at the robot's starting position at about (0m, 0m). The human paths are shown in blue with a blue box at the human's starting position at about (0m, 6.1m). 
In trial configuration labels, L=left, U=unaware/unbiased, and R=right. Shorthand labels (eg. LL, LU) can be read as (robot bias, human action).} 
\label{fig:trajectory_grid}
\end{figure*}

Suppose the robot is unbiased ($b_r=0$)
and approaches a human who is walking straight towards it ($\eta_h = 0$).
In this setting, \eqref{eq:full_z_r} simplifies to
\begin{equation} \label{eq:solo_z_r_reduction}
\dot{z}_r = -d_r z_r + u_r \tanh( \alpha_r z_r).
\end{equation}
The neutral (deadlock) opinion $z_r = 0$ is always an equilibrium solution of \eqref{eq:solo_z_r_reduction}. However, we show that while for small values of attention $u_r$ deadlock is a stable solution, for larger values of $u_r$ it becomes unstable and two symmetric bi-stable solutions emerge corresponding to a strong opinion, one for going left and one for going right. This transition, illustrated in Fig.~\ref{subfig:symmetric_pitchfork} as a plot of equilibrium values of $z_r$ as a function of $u_r$, is called a \textit{pitchfork bifurcation}.

To analyze the deadlock-breaking bifurcation, we linearize the nonlinear opinion equation \eqref{eq:solo_z_r_reduction} around the equilibrium $z_r = 0$ and examine the eigenvalue $\lambda = -d_r + \alpha_r u_r$ of the resulting linearization. 
The sign of $\lambda$ governs the stability of the equilibrium $z_r = 0$. When $\lambda<0$ (resp. $\lambda > 0$), then $z_r=0$, and thus deadlock, is stable (resp. unstable). 

The value of $u_r$ corresponding to $\lambda=0$, computed as $u_r^* = d_r/\alpha_r$, is thus the critical attention value. 
When the robot pays less attention
($u_r< u_r^*$), then $\lambda<0$ and 
the robot remains in deadlock, attempting to go straight to its goal location. However, when the robot pays more attention ($u_r > u_r^*$), $\lambda>0$ and deadlock becomes unstable. For $u_r > u_r^*$ it can be shown that there are two additional symmetric equilibria $z_r^{\rm eq1} = -z_r^{\rm eq2} > 0$ that are both stable. These solutions correspond to a preference for going left ($z_r = z_r^{\rm eq1} > 0$), shown as the positive curve in blue in Fig.~\ref{subfig:symmetric_pitchfork}, and a preference for going right ($z_r = z_r^{\rm eq2} < 0$), shown as the negative curve in blue in Fig.~\ref{subfig:symmetric_pitchfork}. Note that the strength of preferences increases with increasing $u_r>u_r^*$. Because deadlock is unstable, the robot's opinion will necessarily converge on one or the other opinionated solution. Which one it chooses will depend on initial conditions and noise. 



When the robot is biased ($b_r \neq 0$) or the human is approaching the robot obliquely ($\eta_h \neq 0$), the pitchfork bifurcation \textit{unfolds}, as illustrated in Fig.~\ref{subfig:unfolded_pitchfork}. This implies that the robot prefers one side over the other when it passes the human mover.
In particular, it can be shown that the robot prefers to move left if $\gamma_r\tan \eta_h + b_r > 0$, and right if $\gamma_r\tan \eta_h + b_r < 0$.
Also, as we can observe from the diagram in Fig.~\ref{subfig:unfolded_pitchfork}, where the robot has a bias $b_r>0$ for moving left, when $u_r$ becomes sufficiently large, even though the robot favors left, if the robot is already moving right, it continues to move to this side. The analogous holds if $b_r<0$. 

We further illustrate the deadlock-breaking behavior with simulations in Fig.~\ref{subfig:unaware_human_simulations}. The human (trajectory in black) heads straight for the robot. In the unbiased case ($b_r =0$), the robot (trajectory in orange) moves straight just briefly before arbitrarily choosing to go right to pass around the human. This corresponds to behavior indicated by the negative blue curve in Fig.~\ref{subfig:symmetric_pitchfork}. In the biased case ($b_r>0$), the robot (trajectory in purple) follows its bias and moves left, departing even sooner than it did in the unbiased case. This corresponds to the positive blue curve in Fig.~\ref{subfig:unfolded_pitchfork}.

\section{Experiments} \label{section:experiments}


We conducted two laboratory studies with human participants and one wheeled robot, a Clearpath Jackal UGV, moving in the 8m$\times$8m uncluttered space shown in Fig.~\ref{fig:passing_setup}(a). We used a Vicon motion capture system to track the position and orientation of the robot and human movers who wore hats with a set of Vicon markers. The robot used the Vicon data to track the human movers. 
Our experimental goals are threefold: 1) to demonstrate the flexibility of the approach in  that the robot can navigate a space while reliably interacting with multiple human movers in its path over a range of scenarios;
2) to validate the analysis of our algorithm, which shows that the robot is guaranteed to break deadlock, gracefully moving around an oncoming human mover even if the human is unaware of (or ignores) the robot and even if the robot has a bias that conflicts with the passing direction used by the human mover; and 3) to test our hypothesis that the trade-off between more efficient but less reliable passing and less efficient but more reliable passing can be controlled by the single parameter $\beta_r$ in the robot's algorithm \eqref{eq:navi_dynamics}.

\subsection{Validation of Flexibility of the Approach} \label{sec:multi_human_experiments}
\subsubsection{Experimental Setup}
We ran a range of experimental trials each with a different scenario involving the robot and two human participants. In each  trial, the robot and each of the humans were assigned a starting and goal location, which were selected to make the robot and human paths intersect. Human participants could walk along any path at any speed between their starting and goal locations. 

In each trial, the robot was programmed to move at a constant speed of $V_r = 0.75$m/s towards its goal location while adjusting to human movement according to the navigation model \eqref{eq:navi_dynamics} with attention dynamics specified by \eqref{eq:u_r_Hillfxn}.
At any given time, the robot considers only the closest nearby human (according to the measure $\chi/\kappa$) seen within a distance of 20m and an angular range of $(-\frac{\pi}{3}, \frac{\pi}{3})$ with respect to the robot's heading.
If no humans are detected, the robot's attention and opinion are reset to their neutral value, $u_r = z_r = 0$. Results from five representative trials  are shown in Fig.~\ref{fig:two_humans_experiment}. The parameters for \eqref{eq:navi_dynamics} were $d_r = \alpha_r = 0.1$, $\gamma_r = 4$, $k_r = 1.5$, and $\beta_r = \pi/4$. The parameters for \eqref{eq:u_r_Hillfxn} were 
$\underline{u} = 0$ and $R_r = n = 7$. For trials in Fig. \ref{subfig:two_humans_ZU1}-\ref{subfig:two_humans_ZU4}, $\bar{u} = 1.5$ and for the trial in Fig. \ref{subfig:two_humans_ZU5}, $\bar{u} = 2.5$.

\subsubsection{Results}
Fig.~\ref{fig:two_humans_experiment} shows the resulting trajectories and the robot's opinion $z_r$ and attention $u_r$ over the full length of each trial. Temporal markers (dots) are included along the humans' trajectories and the robot's opinion and attention profiles. The top row shows how the robot navigates towards its goal while gracefully modifying its trajectory when encountering humans along its path. The bottom row shows how the robot's attention rises and falls in response to its proximity to a human. When the robot sees a human moving towards its left (resp. right), the opinion becomes negative (resp. positive) and the robot can be observed turning to its right (resp. left). When the robot sees no human to navigate around, its opinion is neutral and its go-to-goal behavior moves the robot towards its goal.

We observe in Fig.~\ref{subfig:two_humans_ZU1}-\ref{subfig:two_humans_ZU3} that the robot's opinion switches sign throughout each trial and that this is reflected in the robot's trajectory, which switches between turns to the left and turns to the right when it passes the human movers. The robot's attention rises and falls as the different participants are seen, maneuvered around, and passed by the robots. 
In Fig.~\ref{subfig:two_humans_ZU4} and \ref{subfig:two_humans_ZU5}, the two human participants approach the robot side-by-side. However, the response of the robot is different in the two cases because the distance between the two participants is different. In  Fig.~\ref{subfig:two_humans_ZU4}, the participants are close together and the robot passes to the right of both, whereas in Fig.~\ref{subfig:two_humans_ZU5}, the participants are further apart, and the robot navigates between them. This is a consequence of the proxy $\tan \eta_h$ that has the same sign for each human mover in the first case but different signs in the second case.

\begin{figure}[t]
\centering
\includegraphics[height=0.38\linewidth]{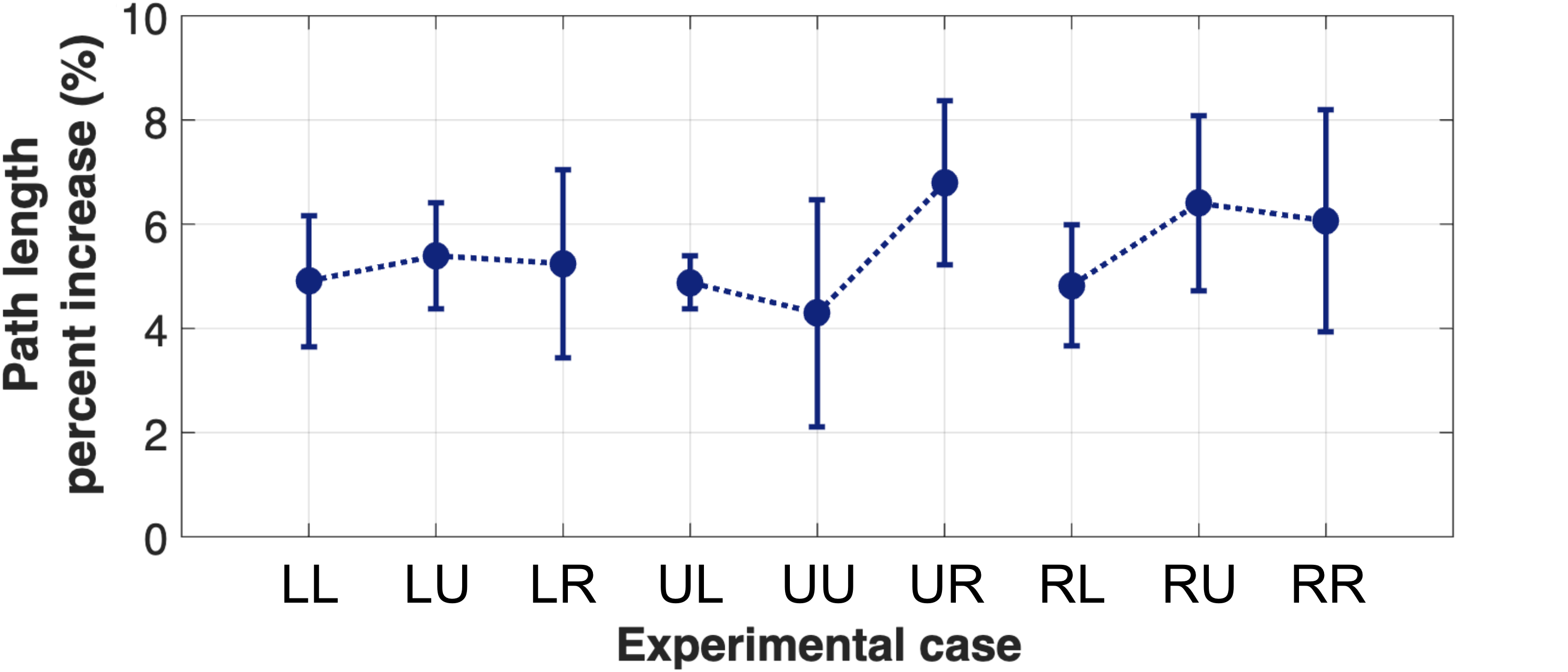}

\caption{Percent increase of the robot's path length for $\beta_r=\pi/4$ compared to $\beta_r=\pi/6$ for each of the nine configurations. Dotted lines link results associated with the same 
robot bias. L/U/R labels as in Fig. \ref{fig:trajectory_grid}.}
\label{fig:path_length_comparison}
\end{figure}

\subsection{Validation of the Deadlock Breaking} \label{sec:one_human_experiments}
\subsubsection{Experimental Setup}
Fixed pairs of starting and goal locations were assigned to the robot and a human participant. The human participant was asked to walk from (0m, 6.1m) to (0m, -1m), and the robot was programmed to navigate from (0m, 0m) to (0m, 6.1m). These locations were selected to make the robot and human move head-on toward one another.

The robot was programmed to move at a constant speed $V_r = 0.7$m/s toward its goal location, modifying its trajectory when encountering movers according to the navigation model \eqref{eq:navi_dynamics} with parameters $d_r = 0.5$, $\alpha_r = 0.1$, $\gamma_r = 3$, $\tau_u = k_r = 1$, and $g(\kappa, \chi; R_r) =\exp(\kappa (R_r - \chi))$ with $R_r = 11$. 
We designed three cases corresponding to three different values of the robot's bias $b_r$: 1) unbiased ($b_r=0$), 2) biased to its left ($b_r=0.5$), and 3) biased to its right ($b_r=-0.5$).

The participant was instructed to walk at their normal pace (their speed was recorded to be $V_h = 1.09 \pm 0.0$3m/s)
towards their goal location according to one of three prompts: 1) go straight,
2) bear to the left, and 3) bear to the right. 

We crossed the three cases for the robot and the three prompts for the human participant for a total of nine different trial configurations. We ran each of these nine different trial configurations five times for a total of 45 trials.
Each of the 45 trials was run with $\beta_r = \pi/4$ and $\beta_r = \pi/6$ in \eqref{eq:navi_dynamics} for a total of 90 trials.



\subsubsection{Results}
Fig.~\ref{fig:trajectory_grid} shows the resultant trajectories of 
the 90 trials organized by configuration on a $3\times3$ grid. For a given configuration and value of $\beta_r$ all five trials are plotted on the same graph. Trials where $\beta_r = \pi/4$ are shaded in yellow and trials where $\beta_r = \pi/6$ are unshaded. It can be observed 
that the robot navigated each trial configuration with similar path structure, regardless of the value of $\beta_r$. 


In all the scenarios where the robot was unbiased (second row of Fig.~\ref{fig:trajectory_grid}), it successfully broke deadlock, verifying the guarantee of deadlock-free navigation provided by the model \eqref{eq:navi_dynamics} and justified in the analysis of Section~\ref{section:analysis}.
In the trials when the human started directly facing the robot and continued walking straight ahead
(UU), as in the simulation Fig.~\ref{subfig:unaware_human_simulations}, the robot quickly formed a strong opinion for one or the other direction. 
The robot chose to go left with about the same frequency that it chose to go right.

\begin{figure}[t]
\centering
\includegraphics[height=0.38\linewidth]{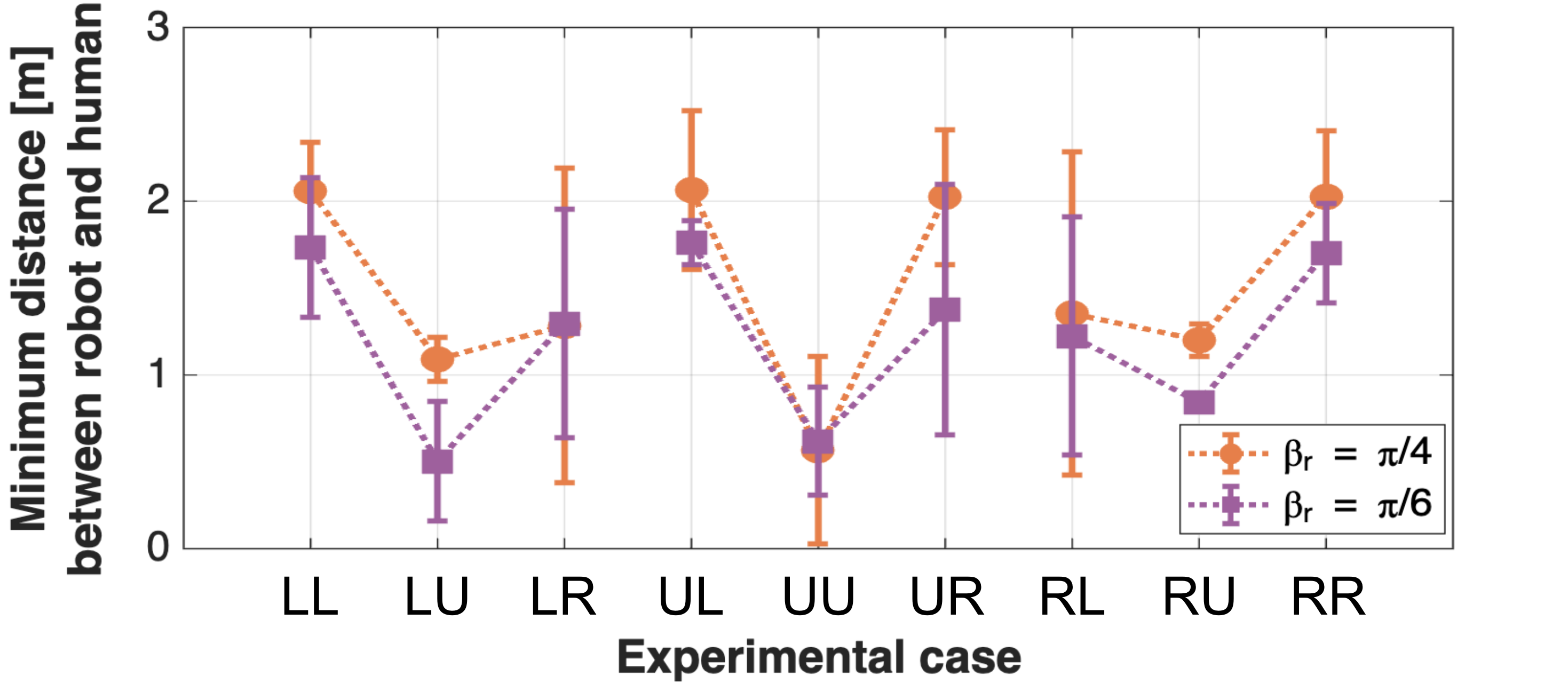}

\caption{Average minimum distance between the robot and human for each of the nine configurations. Dotted lines link results associated with the same $\beta_r$ value (orange line for $\beta_r = \pi/4$, purple line for $\beta_r = \pi/6$) and robot bias. L/U/R labels as in Fig. \ref{fig:trajectory_grid}.}
\label{fig:min_distance_comparison}
\end{figure}

Having a bias allows the robot to rapidly form an initial opinion and break deadlock (turn left if $b_r > 0$ or right if $b_r<0$).   In the scenarios where the robot's bias was in conflict with the action taken by the human ((LR) and (RL) in Fig. \ref{fig:trajectory_grid}), the robot initially moved according to its bias but quickly adapted to the social cues given by the human and passed them 
in a cooperative fashion, i.e., matching the human movement and in opposition to its bias. 
This demonstration of flexibility provides evidence that the robot can reliably adjust its opinion to fit the social context in which it interacts with the human.

The results of Fig.~\ref{fig:trajectory_grid} also provide evidence that a smaller $\beta_r$ (unshaded plots) leads to more efficient (less time to goal) passing around the human as compared to a larger $\beta_r$ (shaded plots). 
Fig.~\ref{fig:path_length_comparison} provides further evidence of the role of $\beta_r$ in tuning efficiency as the percent increase in length of the robot's path for the trials when $\beta_r = \pi/4$ as compared to the case in which $\beta_r = \pi/6$ was uniformly positive, at least 4\% on average. 
Additionally, for each configuration, in trials with larger $\beta_r$ the robot exhibited consistently higher maximum curvature along its path. Trials conducted with $\beta_r = \pi/4$ showed an increase of approximately $22.37\% \pm 6.71\%$ of the maximum curvature of the robot's trajectory as compared to the case $\beta_r = \pi/6$. This confirms that a robot with a larger $\beta_r$ is less efficient. 

Notably, Fig.~\ref{fig:path_length_comparison} shows that the smallest percent increase in robot path length for the increase in $\beta_r$ is in the UU case, when the robot was unbiased and the human unaware of the robot. 
This is consistent with the result that in this trial configuration, the robot took the most time to form a non-neutral opinion and turn to pass the human, which kept its paths in both $\beta_r$ cases closer to the trial space's centerline than observed in other trial configurations. 

Fig. \ref{fig:min_distance_comparison} provides evidence that $\beta_r$ tunes reliability and, together with the results of Fig.~\ref{fig:path_length_comparison}, that $\beta_r$ tunes the efficiency-reliability trade-off, as hypothesized. 
The difference in the minimum distance recorded between the robot and human as they passed one another in each trial configuration for the different $\beta_r$ values is shown in Fig. \ref{fig:min_distance_comparison}. The robot consistently came closer to the human along their paths for $\beta_r = \pi/6$ as compared to $\beta_r=\pi/4$.

For each set of three configurations grouped by the robot's bias, the robot came closest to the human whenever the human was unaware of the robot (i.e. LU, UU, RU). In the other configurations, the robot was able to cooperate with the human to form its opinion and pass the human like the human passed the robot. Without this cooperation, when the robot was the only participant in the passing, the passing distance was consistenly smaller. The minimum distance in the case of the unbiased robot and unaware human was similar for the $\beta_r=\pi/4$ and $\beta_r=\pi/6$ trials. This suggests that this case is the most challenging for the robot independent of $\beta_r$. Still, the general decrease of the minimum distance between the robot and human that comes from a decrease in parameter $\beta_r$ across all other configurations suggests that there is some design threshold where, once passed, the robot could not reliably navigate its way out of collision. Even if the robot's algorithm is such that it can reliably form non-neutral opinions to break deadlock, the design parameters within the model must be sufficiently tuned for use in a real world dynamic context.

\section{Discussion and Final Remarks} \label{section:conclusion}



We present a new proactive approach to social robot navigation that leverages a nonlinear opinion dynamics model 
to enable a robot to rapidly and reliably pass approaching human movers, without requiring a model of human behavior. We show analytically and verify with human-robot experiments that this new navigation algorithm is guaranteed to break deadlock, even when the robot has no bias or evidence from the humans or the environment that one passing direction is better than the other. 
The experiments verify the flexibility of the approach with the robot reliably modifying its trajectory when encountering two human movers in its path.
The experiments also verify that a robot with a bias for passing in one direction can still reliably pass the human mover even if the human chooses to pass in the direction that conflicts with the robot's bias. We show further how design parameters in the robot navigation algorithm can tune the robot's behavior, and verify in the experiments that parameter $\beta_r$ tunes the efficiency-reliability trade-off in the passing problem. 
Future directions include extending the new approach to multi-robot social navigation in more complex scenarios, e.g., with more human movers and more cluttered environments. 
We also plan to investigate an extension that allows for increased sensitivity to changes in context and tuning important trade-offs like efficiency versus reliability. 



\section*{Acknowledgment}
The authors thank Anastasia Bizyaeva and Alessio Franci for 
discussions on 
Section~\ref{section:analysis} and Giovanna Amorim and Justin Lidard for discussions on Section~\ref{section:experiments}.

\bibliographystyle{IEEEtran}
\bibliography{references}

\begin{thebibliography}{10}
\providecommand{\url}[1]{#1}
\csname url@samestyle\endcsname
\providecommand{\newblock}{\relax}
\providecommand{\bibinfo}[2]{#2}
\providecommand{\BIBentrySTDinterwordspacing}{\spaceskip=0pt\relax}
\providecommand{\BIBentryALTinterwordstretchfactor}{4}
\providecommand{\BIBentryALTinterwordspacing}{\spaceskip=\fontdimen2\font plus
\BIBentryALTinterwordstretchfactor\fontdimen3\font minus
  \fontdimen4\font\relax}
\providecommand{\BIBforeignlanguage}[2]{{%
\expandafter\ifx\csname l@#1\endcsname\relax
\typeout{** WARNING: IEEEtran.bst: No hyphenation pattern has been}%
\typeout{** loaded for the language `#1'. Using the pattern for}%
\typeout{** the default language instead.}%
\else
\language=\csname l@#1\endcsname
\fi
#2}}
\providecommand{\BIBdecl}{\relax}
\BIBdecl

\bibitem{ref:Bizyaeva2022_NODM_with_Tunable_Sensitivity}
A.~Bizyaeva, A.~Franci, and N.~E. Leonard, ``Nonlinear opinion dynamics with
  tunable sensitivity,'' \emph{IEEE Trans. on Autom. Control}, vol.~68, no.~3,
  pp. 1415--1430, 2023.

\bibitem{ref:Hamann2018_Opinion_Dynamics_Mobile_Agents}
H.~Hamann, ``\BIBforeignlanguage{en}{Opinion dynamics with mobile agents:
  Contrarian effects by spatial correlations},''
  \emph{\BIBforeignlanguage{en}{Front. Robot. AI}}, vol.~5, p.~63, 2018.

\bibitem{ref:MontesdeOca_2010_Opinion_Dynamics_Decentralized_Decision_Making}
M.~A. Montes~de Oca, E.~Ferrante, N.~Mathews, M.~Birattari, and M.~Dorigo,
  ``Opinion dynamics for decentralized decision-making in a robot swarm,'' in
  \emph{Swarm Intelligence}, 2010, pp. 251--262.

\bibitem{ref:Bizyaeva2022_Decentralized_Control_Switching}
A.~Bizyaeva, G.~Amorim, M.~Santos, A.~Franci, and N.~E. Leonard, ``Switching
  transformations for decentralized control of opinion patterns in signed
  networks: Application to dynamic task allocation,'' \emph{IEEE Control Syst.
  Lett.}, vol.~6, pp. 3463--3468, 2022.

\bibitem{Magrogiannis2023_core_challenges}
C.~Mavrogiannis, F.~Baldini, A.~Wang, D.~Zhao, P.~Trautman, A.~Steinfeld, and
  J.~Oh, ``Core challenges of social robot navigation: A survey,'' \emph{J.
  Hum.-Robot Interact.}, vol.~12, no.~3, pp. 1--39, 2023.

\bibitem{ref:Stone2021_Social_Navigation_Survey}
R.~Mirsky, X.~Xiao, J.~Hart, and P.~Stone, ``Prevention and resolution of
  conflicts in social navigation -- a survey,'' \emph{arXiv:2106.12113}, 2021.

\bibitem{ref:Gao2021_Socially_Aware_Nav_Survey}
Y.~Gao and C.-M. Huang, ``\BIBforeignlanguage{en}{Evaluation of socially-aware
  robot navigation},'' \emph{\BIBforeignlanguage{en}{Front. Robot. AI}},
  vol.~8, art. no. 721317, 2022.

\bibitem{ref:Hart_HumanSignals2020}
J.~Hart, R.~Mirsky, X.~Xiao, S.~Tejeda, B.~Mahajan, J.~Goo, K.~Baldauf,
  S.~Owen, and P.~Stone, ``Using human-inspired signals to disambiguate
  navigational intentions,'' in \emph{Social Robotics}, 2020, pp. 320--331.

\bibitem{ref:Mavrogiannis2022_Social_Momentum}
C.~Mavrogiannis, P.~Alves-Oliveira, W.~Thomason, and R.~A. Knepper, ``Social
  momentum: Design and evaluation of a framework for socially competent robot
  navigation,'' \emph{J. Hum.-Robot Interact.}, vol.~11, no.~2, pp. 1--37,
  2022.

\bibitem{ref:Helbing1995_Social_Force_Model}
D.~Helbing and P.~Moln\'ar, ``Social force model for pedestrian dynamics,''
  \emph{Phys. Rev. E}, vol.~51, pp. 4282--4286, 1995.

\bibitem{ref:Reddy2021_Social_Cues_as_Forces}
A.~K. Reddy, V.~Malviya, and R.~Kala, ``Social cues in the autonomous
  navigation of indoor mobile robots,'' \emph{Int. J. Soc. Robot.}, vol.~13,
  no.~6, pp. 1335--1358, 2021.

\bibitem{ref:Kivrak2021_Human_Inhabited_environments}
H.~Kivrak, F.~Cakmak, H.~Kose, and S.~Yavuz, ``Social navigation framework for
  assistive robots in human inhabited unknown environments,'' \emph{Eng. Sci.
  Technol., Int. J.}, vol.~24, no.~2, pp. 284--298, 2021.

\bibitem{ref:Kirby2009_Companion}
R.~Kirby, R.~Simmons, and J.~Forlizzi, ``Companion: A constraint-optimizing
  method for person-acceptable navigation,'' in \emph{IEEE Int. Symp. Robot
  Human Inter. Commun. (RO-MAN)}, 2009, pp. 607--612.

\bibitem{ref:Samsani2021_Socially_Compliant_Robot_Nav_using_RL}
S.~S. Samsani and M.~S. Muhammad, ``Socially compliant robot navigation in
  crowded environment by human behavior resemblance using deep reinforcement
  learning,'' \emph{IEEE Robot. Autom. Lett (RA-L)}, vol.~6, no.~3, pp.
  5223--5230, 2021.

\bibitem{8202312}
Y.~F. Chen, M.~Everett, M.~Liu, and J.~P. How, ``Socially aware motion planning
  with deep reinforcement learning,'' in \emph{IEEE/RSJ Int. Conf. Robot.
  Intell. Robots Syst. (IROS)}, 2017, pp. 1343--1350.

\bibitem{ref:Kollmitz2020_Learning_from_Physical_Interaction_viaIRL}
M.~Kollmitz, T.~Koller, J.~Boedecker, and W.~Burgard, ``Learning human-aware
  robot navigation from physical interaction via inverse reinforcement
  learning,'' in \emph{IEEE/RSJ Int. Conf. Robot. Intell. Robots Syst. (IROS)},
  2020, pp. 11\,025--11\,031.

\bibitem{ref:Bera2017_SocioSense}
A.~Bera, T.~Randhavane, R.~Prinja, and D.~Manocha, ``Sociosense: Robot
  navigation amongst pedestrians with social and psychological constraints,''
  in \emph{IEEE/RSJ Int. Conf. Robot. Intell. Robots Syst. (IROS)}, 2017, pp.
  7018--7025.

\bibitem{ref:Kretzschmar2016_IRL_Socially_Compliant_Robot}
H.~Kretzschmar, M.~Spies, C.~Sprunk, and W.~Burgard, ``Socially compliant
  mobile robot navigation via inverse reinforcement learning,'' \emph{Int. J.
  Robot. Res.}, vol.~35, no.~11, pp. 1289--1307, 2016.

\bibitem{ref:Okal2016_Learning_SocialNorm_RobotNav_BayesianIRL}
B.~Okal and K.~O. Arras, ``Learning socially normative robot navigation
  behaviors with {B}ayesian inverse reinforcement learning,'' in \emph{IEEE
  Int. Conf. Robot. Autom. (ICRA)}, 2016, pp. 2889--2895.

\bibitem{ref:Che2020_Social_Nav_Explicit_Implicit_Communication}
Y.~Che, A.~M. Okamura, and D.~Sadigh, ``Efficient and trustworthy social
  navigation via explicit and implicit robot-human communication,''
  \emph{{IEEE} Trans. Robotics}, vol.~36, no.~3, pp. 692--707, 2020.

\bibitem{ref:Pacchierotti2005_Hallway_Setting_HRI_Study}
E.~Pacchierotti, H.~Christensen, and P.~Jensfelt, ``Human-robot embodied
  interaction in hallway settings: a pilot user study,'' in \emph{IEEE Int.
  Workshop Robot Human Inter. Commun. (RO-MAN)}, 2005, pp. 164--171.

\bibitem{ref:Thomas2018_Doorway_Negotiation}
J.~Thomas and R.~Vaughan, ``After you: Doorway negotiation for human-robot and
  robot-robot interaction,'' in \emph{IEEE/RSJ Int. Conf. Robot. Intell. Robots
  Syst. (IROS)}, 2018, pp. 3387--3394.

\bibitem{ref:Lu2013_Corridor}
D.~V. Lu and W.~D. Smart, ``Towards more efficient navigation for robots and
  humans,'' in \emph{IEEE/RSJ Int. Conf. Robot. Intell. Robots Syst. (IROS)},
  2013, pp. 1707--1713.

\bibitem{ref:Fiore2013_Robot_Gaze_Proxemic_Behavior}
S.~M. Fiore, T.~J. Wiltshire, E.~J.~C. Lobato, F.~G. Jentsch, W.~H. Huang, and
  B.~Axelrod, ``\BIBforeignlanguage{en}{Toward understanding social cues and
  signals in human-robot interaction: effects of robot gaze and proxemic
  behavior},'' \emph{\BIBforeignlanguage{en}{Front. Psychol.}}, vol.~4, p. 859,
  2013.

\bibitem{ref:Unhelkar2015_HRI_Anticipatory_Indicators}
V.~V. Unhelkar, C.~Pérez-D'Arpino, L.~Stirling, and J.~A. Shah, ``Human-robot
  co-navigation using anticipatory indicators of human walking motion,'' in
  \emph{IEEE Int. Conf. Robot. Autom.(ICRA)}, 2015, pp. 6183--6190.

\bibitem{ref:Ratsamee2015_Face_Orientation_Human_Path_Prediction}
P.~Ratsamee, Y.~Mae, K.~Kamiyama, M.~Horade, M.~Kojima, and T.~Arai, ``Social
  interactive robot navigation based on human intention analysis from face
  orientation and human path prediction,'' \emph{ROBOMECH Journal}, vol.~2,
  art. no. 11, 2015.

\bibitem{ref:Guckenheimer2002}
J.~Guckenheimer and P.~Holmes, \emph{Nonlinear Oscillations, Dynamical Systems,
  and Bifurcations of Vector Fields}, New York: Springer, 2002.

\end{thebibliography}

\end{document}